\documentclass[wcp]{jmlr}

\usepackage{rotating}%
\usepackage{longtable}%

\usepackage{booktabs}
\usepackage[load-configurations=version-1]{siunitx} %
\usepackage{siunitx}
\usepackage{natbib}

\usepackage{times}
\usepackage{latexsym}

\usepackage{bm}
\usepackage{subcaption}
\usepackage{multirow}
\usepackage{mathtools}
\usepackage{url}

\jmlrvolume{101}
\jmlryear{2019}
\jmlrworkshop{ACML 2019}

\title[Cell-aware Stacked LSTMs for Modeling Sentences]{Cell-aware Stacked LSTMs for Modeling Sentences}

\author{}

\author{\Name{Jihun Choi} \Email{jhchoi@europa.snu.ac.kr}\\
    \Name{Taeuk Kim} \Email{taeuk@europa.snu.ac.kr}\\
    \Name{Sang-goo Lee} \Email{sglee@europa.snu.ac.kr}\\
    \addr Department of Computer Science and Engineering, Seoul National University, Seoul, Korea}

\editors{Wee Sun Lee and Taiji Suzuki}

\begin{document}
    
    \maketitle
    
    \begin{abstract}
        We propose a method of stacking multiple long short-term memory (LSTM) layers for modeling sentences.
        In contrast to the conventional stacked LSTMs where only hidden states are fed as input to the next layer, the suggested architecture accepts both hidden and memory cell states of the preceding layer and fuses information from the left and the lower context using the soft gating mechanism of LSTMs.
        Thus the architecture modulates the amount of information to be delivered not only in horizontal recurrence but also in vertical connections, from which useful features extracted from lower layers are effectively conveyed to upper layers.
        We dub this architecture Cell-aware Stacked LSTM (CAS-LSTM) and show from experiments that our models bring significant performance gain over the standard LSTMs on benchmark datasets for natural language inference, paraphrase detection, sentiment classification, and machine translation.
        We also conduct extensive qualitative analysis to understand the internal behavior of the suggested approach.
    \end{abstract}
    
    \section{Introduction}
    \label{sec:intro}
    In the field of natural language processing (NLP), one of the most prevalent neural approaches to obtaining sentence representations is to use recurrent neural networks (RNNs), where words in a sentence are processed in a sequential and recurrent manner.
    Along with their intuitive design, RNNs have shown outstanding performance across various NLP tasks e.g. language modeling \citep{mikolov2010rnnlm,graves2013generating}, machine translation \citep{cho2014nmt,sutskever2014sequence,bahdanau2015nmt}, text classification \citep{zhou2015c,tang2015document}, and parsing \citep{kiperwasser2016parsing,dyer2016rnng}.
    
    Among several variants of the original RNN \citep{elman1990finding}, 
    gated recurrent architectures such as long short-term memory (LSTM) \citep{hochreiter1997long} and gated recurrent unit (GRU) \citep{cho2014nmt} have been accepted as de-facto standard choices for RNNs
    due to their capability of addressing the vanishing and exploding gradient  problem and considering long-term dependencies.
    Gated RNNs achieve these properties by introducing additional gating units that learn to control the amount of information to be transferred or forgotten \citep{goodfellow2016deeplearningboook},
    and are proven to work well without relying on complex optimization algorithms or careful initialization \citep{sutskever2013training}.
    
    Meanwhile, the common practice for further enhancing  the expressiveness of RNNs is to stack multiple RNN layers, each of which has distinct parameter sets (stacked RNN) \citep{schmidhuber1992learning,el1996hierarchical}.
    In stacked RNNs, the hidden states of a layer are fed as input to the subsequent layer, and they are shown to work well due to increased depth \citep{pascanu2014construct} or their ability to capture hierarchical time series \citep{hermans2013training} which are inherent to the nature of the problem being modeled.
    
    \begin{figure}[tb]
        \centering
        \subfigure{%
            \includegraphics[width=0.36\textwidth]{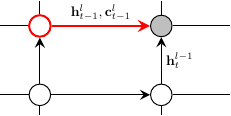}
        }
        \quad
        \subfigure{%
            \includegraphics[width=0.36\textwidth]{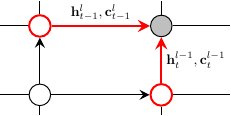}
        }
        \caption{
            Visualization of (\textit{a}) plain stacked LSTM and (\textit{b}) CAS-LSTM.
            The red nodes indicate the blocks whose cell states directly affect the cell state $\mathbf{c}_t^l$.
        }
        \label{fig:comparison}
    \end{figure}
    
    However this setting of stacking RNNs might hinder the possibility of more sophisticated structures since the information from lower layers is simply treated as input to the next layer, rather than as another class of state that participates in core RNN computations.
    Especially for gated RNNs such as LSTMs and GRUs, this means that the vertical layer-to-layer connections cannot fully benefit from the carefully constructed gating mechanism used in temporal transitions.
    
    In this paper, we study a method of constructing multi-layer LSTMs where memory cell states from the previous layer are used in controlling the vertical information flow.
    This system utilizes states from the left and the lower context equally in computation of the new state, thus the information from lower layers is elaborately filtered and reflected through a soft gating mechanism. %
    Our method is easy-to-implement, effective, and can replace conventional stacked LSTMs without much modification of the overall architecture.
    
    We call this architecture Cell-aware Stacked LSTM, or CAS-LSTM, and evaluate our method on multiple benchmark tasks: natural language inference, paraphrase identification, sentiment classification, and machine translation.
    From experiments we show that the CAS-LSTMs consistently outperform typical stacked LSTMs, opening the possibility of performance improvement of architectures based on stacked LSTMs.
    
    Our contribution is summarized as follows.
    Firstly, we bring the idea of utilizing states coming from multiple directions to construction of stacked LSTM and apply the idea to the research of sentence representation learning.
    There is some prior work addressing the idea of incorporating more than one type of state \citep{graves2007mdrnn,kalchbrenner2016grid,zhang2016highway}, however to the best of our knowledge there is little work on applying the idea to modeling sentences for better understanding of natural language text.
    
    Secondly, we conduct extensive evaluation of the proposed method and empirically prove its effectiveness.
    The CAS-LSTM architecture provides consistent performance gains over the stacked LSTM in all benchmark tasks: natural language inference, paraphrase identification, sentiment classification, and machine translation.
    Especially in SNLI, SST-2, and Quora Question Pairs datasets, our models outperform or at least are on par with the state-of-the-art models.
    We also conduct thorough qualitative analysis to understand the dynamics of the suggested approach.
    
    This paper is organized in the following way.
    We study prior work related to our objective in \S\ref{sec:related}, and
    \S\ref{sec:caslstm} gives a detailed description about the proposed method.
    Experimental results are given in \S\ref{sec:experiments}, and 
    \S\ref{sec:conclusion} concludes this paper.
    
    \section{Related Work}
    \label{sec:related}
    In this section, we summarize prior work related to the proposed method.
    We group the previous work that motivated our work into three classes: i) enhancing interaction between vertical layers, ii) RNN architectures that accepts latticed data, and iii) tree-structured RNNs.
    
    \paragraph{Stacked RNNs.}
    There is some prior work on methods of stacking RNNs beyond the plain stacked RNNs \citep{schmidhuber1992learning,el1996hierarchical}.
    Residual LSTMs \citep{kim2017residual,tran2017stack} add residual connections between the hidden states computed at each LSTM layer, and shortcut-stacked LSTMs \citep{nie2017shortcut} concatenate hidden states from all previous layers to make the backpropagation path short.
    In our method, the lower context is aggregated via a gating mechanism, and we believe it modulates the amount of information to be transmitted in a more efficient and effective way than vector addition or concatenation.
    Also, compared to concatenation, our method does not significantly increase the number of parameters.\footnote{The $l$-th layer of a typical stacked LSTM requires $(d_{l-1} + d_l + 1) \times 4d_l$ parameters, and the $l$-th layer of a shortcut-stacked LSTM requires $(\sum_{k=0}^{l-1} {d_k} + d_l + 1) \times 4d_l$ parameters. CAS-LSTM uses $(d_{l-1} + d_l + 1) \times 5d_l$ parameters at the $l$-th ($l>1$) layer.}
    
    Highway LSTMs \citep{zhang2016highway} and depth-gated LSTMs \citep{yao2015depth} are similar to our proposed models in that they use cell states from the previous layer, and they are successfully applied to the field of automatic speech recognition and language modeling.
    However in contrast to CAS-LSTM, where the additional forget gate aggregates the previous layer states and thus contexts from the left and below participate in computation equitably, in Highway LSTMs and depth-gated LSTMs the states from the previous time step are not considered in computing vertical gates.
    The comparison of our method and this architecture is presented in \S\ref{exp:variations}.
    
    \paragraph{Multidimensional RNNs.}
    There is another line of research that aims to extend RNNs to operate with multidimensional inputs.
    Grid LSTMs \citep{kalchbrenner2016grid} are a general $n$-dimensional LSTM architecture that accepts $n$ sets of hidden and cell states as input and yields $n$ sets of states as output, in contrast to our architecture, which emits a single set of states.
    In their work, the authors utilize 2D and 3D Grid LSTMs in character-level language modeling and machine translation respectively and achieve performance improvement.
    Multidimensional RNNs \citep{graves2007mdrnn,graves2009offline} have similar formulation to ours, except that they reflect cell states via simple summation and weights for all columns (vertical layers in our case) are tied.
    However they are only employed to model multidimensional data such as images of handwritten text with RNNs, rather than stacking RNN layers for modeling sequential data.
    From this view, CAS-LSTM could be interpreted as an extension of two-dimensional LSTM architecture that accepts a 2D input $\{\mathbf{h}_t^l\}_{t=1,l=0}^{T,L}$ where $\mathbf{h}_t^l$ represents the hidden state at time $t$ and layer $l$.
    
    \paragraph{Tree-structured RNNs.}
    The idea of having multiple states is also related to tree-structured RNNs \citep{goller1996learning,socher2011parsing}.
    Among them, tree-structured LSTMs (tree-LSTMs) \citep{tai2015treelstm,zhu2015treelstm,le2015treelstm} are similar to ours in that they use both hidden and cell states of children nodes.
    In tree-LSTMs, states of children nodes are regarded as input, and they participate in computing the states of a parent node equally through weight-shared or weight-unshared projection.
    From this perspective, each CAS-LSTM layer can be seen as a binary tree-LSTM where the structures it operates on are fixed to right-branching trees.
    
    Indeed, our work is motivated by the recent analysis \citep{williams2018do,shi2018ontree} on latent tree learning models \citep{yogatama2017learning,choi2018learning} which has shown that tree-LSTM models outperform the sequential LSTM models even when the resulting parsing strategy generates strictly left- or right-branching parses, where a tree-LSTM model should read words in the manner identical to a sequential LSTM model.
    We argue that the active use of cell state in computation could be one reason of these counter-intuitive results and empirically prove the hypothesis in this work.

    \section{Model Description}
    \label{sec:caslstm}
    
    \begin{figure}[tb]
        \centering
        \includegraphics[width=0.5\textwidth]{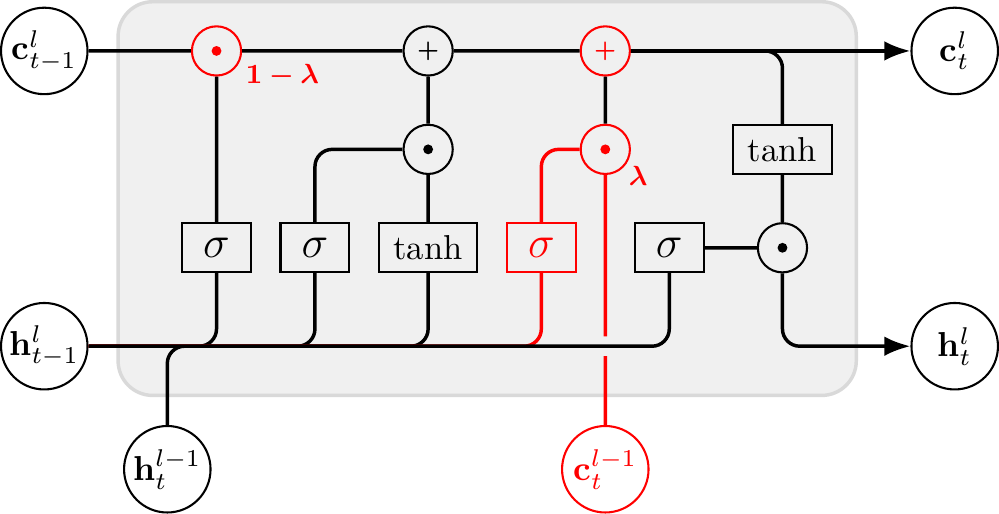}
        \caption{Schematic diagram of a CAS-LSTM block.}
        \label{fig:diagram}
    \end{figure}
    
    In this section, we give the detailed formulation of architectures used in experiments.
    \subsection{Stacked LSTMs}
    \label{ssec:stacked-lstm}
    While there exist various versions of LSTM formulation, in this work we use the following, the most common variant:
    \begin{align}
    \mathbf{i}_t^l &= \sigma(\mathbf{W}_i^l \mathbf{h}_t^{l-1} + \mathbf{U}_i^l \mathbf{h}_{t-1}^l + \mathbf{b}_i^l) \\
    \mathbf{f}_t^l &= \sigma(\mathbf{W}_f^l \mathbf{h}_t^{l-1} + \mathbf{U}_f^l \mathbf{h}_{t-1}^l + \mathbf{b}_f^l) \\
    \tilde{\mathbf{c}}_t^l &= \tanh(\mathbf{W}_c^l \mathbf{h}_t^{l-1} + \mathbf{U}_c^l \mathbf{h}_{t-1}^l + \mathbf{b}_c^l) \\
    \mathbf{o}_t^l &= \sigma(\mathbf{W}_o^l \mathbf{h}_t^{l-1} + \mathbf{U}_o^l \mathbf{h}_{t-1}^l + \mathbf{b}_o^l) \\
    \mathbf{c}_t^l &= \mathbf{i}_t^l \odot \tilde{\mathbf{c}}_t^l + \mathbf{f}_t^l \odot \mathbf{c}_{t-1}^l \label{eq:conventional-cell}\\
    \mathbf{h}_t^l &= \mathbf{o}_t^l \odot \tanh(\mathbf{c}_t^l),
    \end{align}
    where $t \in \{1,\cdots,T\}$ and $l \in \{1,\cdots,L\}$. $\mathbf{W}_{\cdot}^{l} \in \mathbb{R}^{d_l \times d_{l-1}}$, $\mathbf{U}_{\cdot}^{l} \in \mathbb{R}^{d_l \times d_l}$, $\mathbf{b}_{\cdot}^{l} \in \mathbb{R}^{d_l}$ are trainable parameters,
    and $\sigma(\cdot)$ and $\tanh(\cdot)$ are the sigmoid and the hyperbolic tangent function respectively.
    Also we assume that $\mathbf{h}_t^0=\mathbf{x}_t \in \mathbb{R}^{d_0}$ where $\mathbf{x}_t$ is the $t$-th element of an input sequence.
    
    The input gate $\mathbf{i}_t^l$ and the forget gate $\mathbf{f}_t^l$ control the amount of information transmitted from $\tilde{\mathbf{c}}_t^l$ and $\mathbf{c}_{t-1}^l$, the candidate cell state and the previous cell state, to the new cell state $\mathbf{c}_t^l$.
    Similarly the output gate $\mathbf{o}_t^l$ soft-selects which portion of the cell state $\mathbf{c}_t^l$ is to be used in the final hidden state.
    
    We can clearly see that the cell states $\mathbf{c}_{t-1}^l$, $\tilde{\mathbf{c}}_t^l$, $\mathbf{c}_t^l$ play a crucial role in forming horizontal recurrence.
    However the current formulation does not consider the cell state from $(l-1)$-th layer ($\mathbf{c}_t^{l-1}$) in computation and thus the lower context is reflected only through the rudimentary way, hindering the possibility of controlling vertical information flow.
    
    \subsection{Cell-aware Stacked LSTMs}
    Now we extend the stacked LSTM formulation defined above to address the problem noted in the previous subsection.
    To enhance the interaction between layers in a way similar to how LSTMs keep and forget the information from the previous time step, we introduce the \textit{additional forget gate} $\mathbf{g}_t^l$ that determines whether to accept or ignore the signals coming from the previous layer.
    
    The proposed Cell-aware Stacked LSTM (CAS-LSTM) architecture is defined as follows:
    \begin{align}
    \mathbf{i}_t^l &= \sigma(\mathbf{W}_i^l \mathbf{h}_t^{l-1} + \mathbf{U}_i^l \mathbf{h}_{t-1}^l + \mathbf{b}_i^l) \\
    \mathbf{f}_t^l &= \sigma(\mathbf{W}_f^l \mathbf{h}_t^{l-1} + \mathbf{U}_f^l \mathbf{h}_{t-1}^l + \mathbf{b}_f^l) \\
    \mathbf{g}_t^l &= \sigma(\mathbf{W}_g^l \mathbf{h}_t^{l-1} + \mathbf{U}_g^l \mathbf{h}_{t-1}^l + \mathbf{b}_g^l) \label{eq:new-forget-gate}\\
    \tilde{\mathbf{c}}_t^l &= \tanh(\mathbf{W}_c^l \mathbf{h}_t^{l-1} + \mathbf{U}_c^l \mathbf{h}_{t-1}^l + \mathbf{b}_c^l) \\
    \mathbf{o}_t^l &= \sigma(\mathbf{W}_o^l \mathbf{h}_t^{l-1} + \mathbf{U}_o^l \mathbf{h}_{t-1}^l + \mathbf{b}_o^l) \\
    \mathbf{c}_t^l &= \mathbf{i}_t^l \odot {\tilde{\mathbf{c}}}_t^l + (\bm{1 - \lambda})\odot\mathbf{f}_t^l \odot \mathbf{c}_{t-1}^l + \bm{\lambda}\odot\mathbf{g}_t^l \odot \mathbf{c}_t^{l-1} \\
    \mathbf{h}_t^l &= \mathbf{o}_t^l \odot \tanh(\mathbf{c}_t^l),
    \end{align}
    where $l > 1$ and $d_l=d_{l-1}$.
    $\bm\lambda$ can either be a vector of constants or parameters.
    When $l=1$, the equations defined in the previous subsection are used.
    Therefore, it can be said that each non-bottom layer of CAS-LSTM accepts two sets of hidden and cell states---one from the left context and the other from the below context.
    The left and the below context participate in computation with the equivalent procedure so that the information from lower layers can be efficiently propagated.
    Fig. \ref{fig:comparison} compares CAS-LSTM to the conventional stacked LSTM architecture,
    and Fig. \ref{fig:diagram} depicts the computation flow of the CAS-LSTM.
    
    We argue that considering $\mathbf{c}_t^{l-1}$ in computation is beneficial for the following reasons.
    First, contrary to $\mathbf{h}_t^{l-1}$, $\mathbf{c}_t^{l-1}$ contains information which is not filtered by $\mathbf{o}_t^{l-1}$.
    Thus a model that directly uses $\mathbf{c}_t^{l-1}$ does not rely solely on $\mathbf{o}_t^{l-1}$ for extracting information, due to the fact that it has access to the raw information $\mathbf{c}_t^{l-1}$, as in temporal connections.
    In other words, $\mathbf{o}_t^{l-1}$ no longer has to take all responsibility for selecting useful features for both horizontal and vertical transitions, and the burden of selecting information is shared with $\mathbf{g}_t^l$.
    
    \begin{figure}
        \centering
        \includegraphics[width=0.42\textwidth]{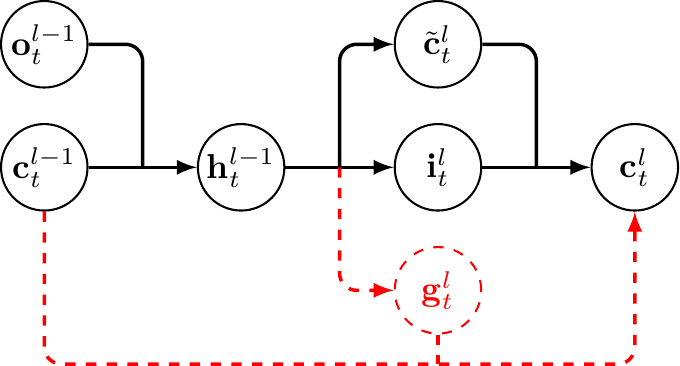}
        \caption{
            Visualization of paths between $\mathbf{c}_t^{l-1}$ and $\mathbf{c}_t^l$.
            In CAS-LSTM, the direct connection between $\mathbf{c}_t^{l-1}$ and $\mathbf{c}_t^l$ exists (denoted as red dashed lines).
        }
        \label{fig:ca_path}
    \end{figure}
    
    Another advantage of using the $\mathbf{c}_t^{l-1}$ lies in the fact that it directly connects $\mathbf{c}_t^{l-1}$ and $\mathbf{c}_t^l$.
    This direct connection could help and stabilize training, since the terminal error signals can be easily backpropagated to the model parameters by the shortened propagation path.
    Fig. \ref{fig:ca_path} illustrates paths between the two cell states.
    
    Regarding $\bm\lambda$, we find experimentally that there is little difference between having it be a constant and a trainable vector bounded in $(0, 1)$, and we practically find that setting $\lambda_i=0.5$ works well across multiple experiments.
    We also experimented with the architecture without $\bm\lambda$ i.e. two cell states are combined by unweighted summation similar to multidimensional RNNs \citep{graves2009offline}, and found that it leads to performance degradation and unstable convergence, likely due to mismatch in the range of cell state values between layers ($(-2, 2)$ for the first layer and $(-3, 3)$ for the others).
    Experimental results on various $\bm\lambda$ are presented in \S\ref{exp:variations}.

    \subsection{Sentence Encoders}
    For text classification tasks, a variable-length sentence should be represented as a fixed-length vector.
    We describe the sentence encoder architectures used in experiments in this subsection.
    
    First, we assume that a sequence of $T$ one-hot word vectors is given as input: $(\mathbf{w}_1, \cdots, \mathbf{w}_T)$, $\mathbf{w}_t \in \mathbb{R}^{|V|}$ where $V$ is the vocabulary set.
    The words are projected to corresponding word representations: $\mathbf{X}=(\mathbf{x}_1, \cdots, \mathbf{x}_T)$ where $\mathbf{x}_{t} = \mathbf{E}^\top \mathbf{w}_t \in \mathbb{R}^{d_0}$, $\mathbf{E}\in \mathbf{R}^{|V| \times d_0}$.
    Then $\mathbf{X}$ is fed to a $L$-layer CAS-LSTM model, resulting in the representations $\mathbf{H}=(\mathbf{h}_1^L, \cdots, \mathbf{h}_T^L)\in \mathbb{R}^{T\times d_L}$.
    The encoded sentence representation $\mathbf{s} \in \mathbb{R}^{d_L}$ is computed by max-pooling $\mathbf{H}$ over time as in the work of \citet{conneau2017infersent}.
    Similar to their results, from preliminary experiments we found that the max-pooling performs consistently better than the mean-pooling and the last-pooling.
    
    For better modeling of semantics, a bidirectional CAS-LSTM network may also be used.
    In the bidirectional case, the representations obtained by left-to-right reading $\mathbf{H}=(\mathbf{h}_1^L, \cdots, \mathbf{h}_T^L) \in \mathbb{R}^{T\times d_L}$
    and those by right-to-left reading $\widehat{\mathbf{H}}=(\widehat{\mathbf{h}}_1^L, \cdots, \widehat{\mathbf{h}}_T^L) \in \mathbb{R}^{T \times d_L}$ are concatenated and max-pooled to yield the sentence representation $\mathbf{s} \in \mathbb{R}^{2d_L}$.
    We call this bidirectional architecture Bi-CAS-LSTM in experiments.
    
    To predict the final task-specific label, we apply a task-specific feature extraction function $\phi$ to the sentence representation(s) and feed the extracted features to a classifier network.
    For the classifier network, a multi-layer perceptron (MLP) with the ReLU activation followed by the linear projection and the softmax function is used:
    \begin{equation}
    P(\mathbf{y}|\mathbf{X}) = \text{softmax}(\mathbf{W}_c\text{MLP}(\phi(\cdot))),
    \end{equation}
    where $\mathbf{W}_c \in \mathbb{R}^{|L|\times d_h}$, $|L|$ is the number of label classes, and $d_h$ the dimension of the MLP output.

    \section{Experiments}
    \label{sec:experiments}
    We evaluate our method on three benchmark tasks on sentence encoding: natural language inference (NLI), paraphrase identification (PI), and sentiment classification.
    To further demonstrate the general applicability of our method on text generation, we also evaluate the proposed method on machine translation.
    In addition, we conduct analysis on gate values model variations for the understanding of the architecture.
    We refer readers to the supplemental material for detailed experimental settings.
    The code will be made public for reproduction.
    
    For the NLI and PI tasks, there exists architectures specializing in sentence pair classification.
    However in this work we confine our model to the architecture that encodes each sentence using a shared encoder without any inter-sentence interaction, in order to focus on the effectiveness of the architectures in extracting semantics.
    But note that the applicability of CAS-LSTM is not limited to sentence encoder--based approaches.
    
    \subsection{Natural Language Inference}
    
    \begin{table*}[t]
        \centering
        \begin{tabular}{l r r}
            \hline
            \bf{Model} & \bf{Acc. (\%)} & \bf {\# Params} \\
            \hline
            300D LSTM \citep{bowman2016spinn} & 80.6 & 3.0M \\
            300D TBCNN \citep{mou2016snli} & 82.1 & 3.5M \\
            300D SPINN-PI \citep{bowman2016spinn} & 83.2 & 3.7M \\
            600D BiLSTM with intra-attention \citep{liu2016learning} & 84.2 & 2.8M \\
            4096D BiLSTM with max-pooling \citep{conneau2017infersent} & 84.5 & 40M \\
            300D BiLSTM with gated pooling \citep{chen2017gated} & 85.5 & 12M \\
            300D Gumbel Tree-LSTM \citep{choi2018learning} & 85.6 & 2.9M \\
            600D Shortcut stacked BiLSTM \citep{nie2017shortcut} & 86.1 & 140M \\
            300D Reinforced self-attention network \citep{shen2018reinforced} & 86.3 & 3.1M \\
            600D BiLSTM with generalized pooling \citep{chen2018generalized} & 86.6 & 65M \\
            \hline
            300D 2-layer CAS-LSTM (ours) & 86.4 & 2.9M \\
            300D 2-layer Bi-CAS-LSTM (ours) & {86.8} & 6.8M \\
            300D 3-layer CAS-LSTM (ours) & 86.4 & 4.8M  \\
            300D 3-layer Bi-CAS-LSTM (ours) & \bf{87.0} & 8.6M \\
            \hline
        \end{tabular}
        \caption{Results of the models on the SNLI dataset.}
        \label{table:snli}
    \end{table*}
    
    \begin{table*}[t]
        \centering
        \begin{tabular}{l r r r}
            \hline
            \bf{Model} & \bf{In (\%)} & \bf{Cross (\%)} & \bf{\# Params} \\
            \hline
            CBOW \citep{williams2018mnli} & 64.8 & 64.5 & - \\
            BiLSTM \citep{williams2018mnli} & 66.9 & 66.9 & - \\
            Shortcut stacked BiLSTM \citep{nie2017shortcut}$^\ast$ & \bf{74.6} & 73.6 & 140M \\
            BiLSTM with gated pooling \citep{chen2017gated} & 73.5 & 73.6 & 12M \\
            BiLSTM with generalized pooling \citep{chen2018generalized} & 73.8 & \bf{74.0} & 18M$^{\ast\ast}$ \\
            \hline
            2-layer CAS-LSTM (ours) & 74.0 & 73.3 & 2.9M \\
            2-layer Bi-CAS-LSTM (ours) & \bf{74.6} & 73.7 & 6.8M \\
            3-layer CAS-LSTM (ours) & 73.8 & 73.1 & 4.8M \\
            3-layer Bi-CAS-LSTM (ours) & 74.2 & 73.4 & 8.6M \\
            \hline
        \end{tabular}
        \caption{Results of the models on the MultiNLI dataset.
            `In' and `Cross' represent accuracy calculated from the matched and mismatched test set respectively.
            $^\ast$: SNLI dataset is used as additional training data. $^{\ast\ast}$: computed from hyperparameters provided by the authors.}
        \label{table:mnli}
    \end{table*}
    
    For the evaluation of performance of the proposed method on the NLI task, SNLI \citep{bowman2015snli} and MultiNLI \citep{williams2018mnli} datasets are used.
    The objective of both datasets is to predict the relationship between a premise and a hypothesis sentence: \textit{entailment}, \textit{contradiction}, and \textit{neutral}.
    SNLI and MultiNLI datasets are composed of about 570k and 430k premise-hypothesis pairs respectively.
    
    GloVe pretrained word embeddings\footnote{\url{https://nlp.stanford.edu/projects/glove/}} \citep{pennington2014glove} are used and remain fixed during training.
    The dimension of encoder states ($d_l$) is set to 300 and a 1024D MLP with one or two hidden layers is used.
    We apply dropout \citep{srivastava2014dropout} to the word embeddings and the MLP layers.
    The features used as input to the MLP classifier are extracted by the following equation:
    \begin{equation}
    \label{eq:nli-matching}
    \phi(\mathbf{s}_1, \mathbf{s}_2) = \mathbf{s}_1 \oplus \mathbf{s}_2 \oplus |\mathbf{s}_1 - \mathbf{s}_2| \oplus (\mathbf{s}_1 \odot \mathbf{s}_2),
    \end{equation}
    where $\oplus$ is the vector concatenation operator.
    
    Table \ref{table:snli} and \ref{table:mnli} contain results of the models on SNLI and MultiNLI datasets.
    Along with other state-of-the-art models, the tables include several stacked LSTM--based models to facilitate comparison of our work with prior related work.
    \citet{liu2016learning,chen2017gated,chen2018generalized} adopt advanced pooling algorithms motivated by the attention mechanism to obtain a fixed-length sentence vector.
    \citet{nie2017shortcut} use the concatenation of all outputs from previous layers as input to the next layer.
    
    In SNLI, our best model achieves the accuracy of 87.0\%, which is the new state-of-the-art among the sentence encoder--based models, with relatively fewer parameters.
    Similarly in MultiNLI, our models match the accuracy of state-of-the-art models in both in-domain (matched) and cross-domain (mismatched) test sets.
    Note that only the GloVe word vectors are used as word representations, as opposed to some models that introduce character-level features.
    It is also notable that our proposed architecture does not restrict the selection of pooling method; the performance could further be improved by replacing max-pooling with other advanced algorithms e.g. intra-sentence attention \citep{liu2016learning} and generalized pooling \citep{chen2018generalized}.
    
    \subsection{Paraphrase Identification}
    
    \begin{table}[t]
        \centering
        \begin{tabular}{l r}
            \hline
            \bf{Model} & \bf{Acc. (\%)} \\
            \hline
            CNN \citep{wang2017bilateral} & 79.6 \\
            LSTM \citep{wang2017bilateral} & 82.6 \\
            Multi-Perspective LSTM \citep{wang2017bilateral} & 83.2 \\
            LSTM + ElBiS \citep{choi2018elbis} & 87.3 \\
            REGMAPR (BASE+REG) \citep{brahma2018regmapr} & 88.0 \\
            \hline
            CAS-LSTM (ours) &  88.4 \\
            Bi-CAS-LSTM (ours) & \bf{88.6} \\
            \hline 
        \end{tabular}
        \caption{Results of the models on the Quora Question Pairs dataset.}
        \label{table:quora}
    \end{table}
    
    We use Quora Question Pairs dataset \citep{wang2017bilateral} in evaluating the performance of our method on the PI task.
    The dataset consists of over 400k question pairs, and each pair is annotated with whether the two sentences are paraphrase of each other or not.
    
    Similarly to the NLI experiments, GloVe pretrained vectors, 300D encoders, and 1024D MLP are used.
    The number of CAS-LSTM layers is fixed to 2 in PI experiments.
    Two sentence vectors are aggregated using the following equation and fed as input to the classifier.
    \begin{equation}
    \label{eq:pi-matching}
    \phi(\mathbf{s}_1, \mathbf{s}_2) = |\mathbf{s}_1 - \mathbf{s}_2| \oplus (\mathbf{s}_1 \odot \mathbf{s}_2)
    \end{equation}
    
    The results on the Quora Question Pairs dataset are summarized in Table \ref{table:quora}.
    Again we can see that our models outperform other models, especially compared to conventional LSTM--based models.
    Also note that Multi-Perspective LSTM \citep{wang2017bilateral}, LSTM + ElBiS \citep{choi2018elbis}, and REGMAPR (BASE+REG) \citep{brahma2018regmapr} in Table \ref{table:quora} are approaches that focus on designing a more sophisticated function for aggregating two sentence vectors, and their aggregation functions could be also applied to our work for further improvement.
    
    \subsection{Sentiment Classification}
    
    \begin{table*}[t]
        \centering
        \begin{tabular}{l r r}
            \hline
            \bf{Model} & \bf{SST-2 (\%)} & \bf{SST-5 (\%)} \\
            \hline
            Recursive Neural Tensor Network \citep{socher2013recursive} & 85.4 & 45.7 \\
            2-layer LSTM \citep{tai2015treelstm} & 86.3 & 46.0 \\
            2-layer BiLSTM \citep{tai2015treelstm} & 87.2 & 48.5 \\
            Constituency Tree-LSTM \citep{tai2015treelstm} & 88.0 & 51.0 \\
            Constituency Tree-LSTM with recurrent dropout \citep{looks2017deep} & 89.4 & 52.3 \\
            byte mLSTM \citep{radford2017learning}$^\ast$ & \underline{91.8} & 52.9 \\
            Gumbel Tree-LSTM \citep{choi2018learning} & 90.7 & \bf{53.7} \\
            BCN + Char + ELMo \citep{peters2018elmo}$^\ast$ & - & \underline{54.7} \\
            \hline
            2-layer CAS-LSTM (ours) & 91.1 & 53.0	 \\
            2-layer Bi-CAS-LSTM (ours) & \bf{91.3} & 53.6 \\
            \hline
        \end{tabular}
        \caption{Results of the models on the SST dataset. $^\ast$: models pretrained on large external corpora are used.}
        \label{table:sst}
    \end{table*}
    
    In evaluating sentiment classification performance, the Stanford Sentiment Treebank (SST) \citep{socher2013recursive} is used.
    It consists of about 12,000 binary-parsed sentences where constituents (phrases) of each parse tree are annotated with a sentiment label (\textit{very positive}, \textit{positive}, \textit{neutral}, \textit{negative}, \textit{very negative}).
    Following the convention of prior work, all phrases and their labels are used in training but only the sentence-level data are used in evaluation.
    
    In evaluation we consider two settings, namely SST-2 and SST-5, the two differing only in their level of granularity with regard to labels.
    In SST-2, data samples annotated with `neutral' are ignored from training and evaluation.
    The two positive labels (very positive, positive) are considered as the same label, and similarly for the two negative labels.
    As a result 98,794/872/1,821 data samples are used in training/validation/test, and the task is considered as a binary classification problem.
    In SST-5, all 318,582/1,101/2,210 data samples are used and the task is a 5-class classification problem.
    
    Since the task is a single-sentence classification problem, we use the sentence representation itself as input to the classifier.
    We use 300D GloVe vectors, 2-layer 150D or 300D encoders, and a 300D MLP classifier for the models,
    however unlike previous experiments we tune the word embeddings during training.
    The results on SST are listed in Table \ref{table:sst}.
    Our models clearly outperform plain LSTM- and BiLSTM-based models, and are competitive to other state-of-the-art models, without utilizing parse tree information.
    
    \subsection{Machine Translation}
    \begin{table}[tb]
        \centering
        \begin{tabular}{l l}
            \hline
            \bf{Model} & \bf{BLEU} \\
            \hline
            256D LSTM & 28.1 $\pm$ 0.22 \\
            256D CAS-LSTM & 28.8 $\pm$ 0.04$^\ast$ \\
            247D CAS-LSTM & 28.7 $\pm$ 0.07$^\ast$ \\
            \hline
        \end{tabular}
        \caption{
            Results of the models on the IWSLT 2014 de-en dataset.
            $^\ast$: $p < 0.0005$ (one-tailed paired t-test).
        }
        \label{table:mt}
    \end{table}
    We use the IWSLT 2014 machine evaluation campaign dataset \citep{cettolo2014report} in machine translation experiments.
    We used the fairseq library\footnote{\url{https://github.com/pytorch/fairseq}} \citep{gehring2017fairseq} for experiments.
    Moses tokenizer\footnote{\url{https://github.com/moses-smt/mosesdecoder/blob/master/scripts/tokenizer/tokenizer.perl}} is used for word tokenization and the byte pair encoding \citep{sennrich2016bpe} is applied to confine the size of the vocabulary set up to 10,000.
    
    Similar to \citet{wiseman2016nmt}, a 2-layer 256D sequence-to-sequence LSTM model with the attentional decoder is used as baseline, and we replace the encoder and the decoder network with the proposed architecture for the evaluation of performance improvement.
    For decoding, beam search with $B=10$ is used.
    For fair comparison, we tune hyperparameters for all models based on the performance on the validation dataset and train the same model for five times with different random seeds.
    Also, to cancel out the increased number of parameters, we experiment with the 247D CAS-LSTM model which has the roughly same number of parameters as the baseline model (8.2M).
    
    From Table \ref{table:mt}, we can see that the CAS-LSTM models bring significant performance gains over the baseline model.
    
    \subsection{Forget Gate Analysis}
    
    \begin{figure}[t]
        \centering
        \subfigure{%
            \label{fig:gate-a}
            \includegraphics[width=0.3\textwidth]{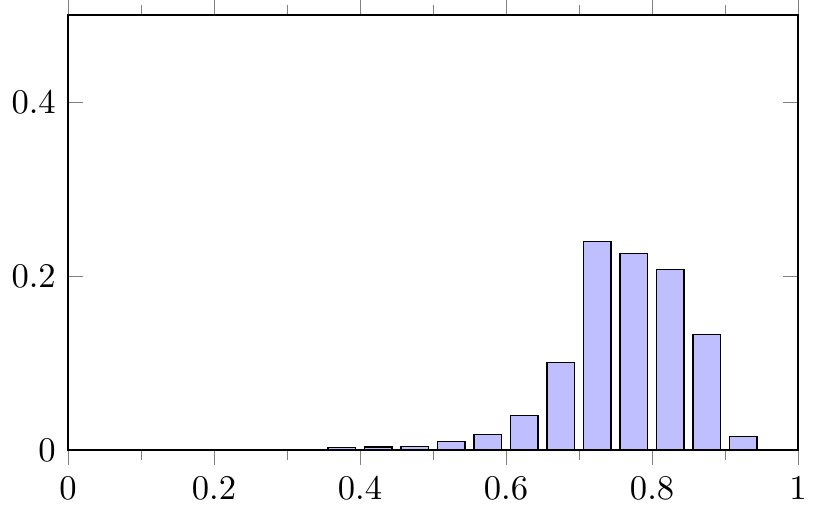}
        }
        \quad
        \subfigure{%
            \label{fig:gate-b}
            \includegraphics[width=0.3\textwidth]{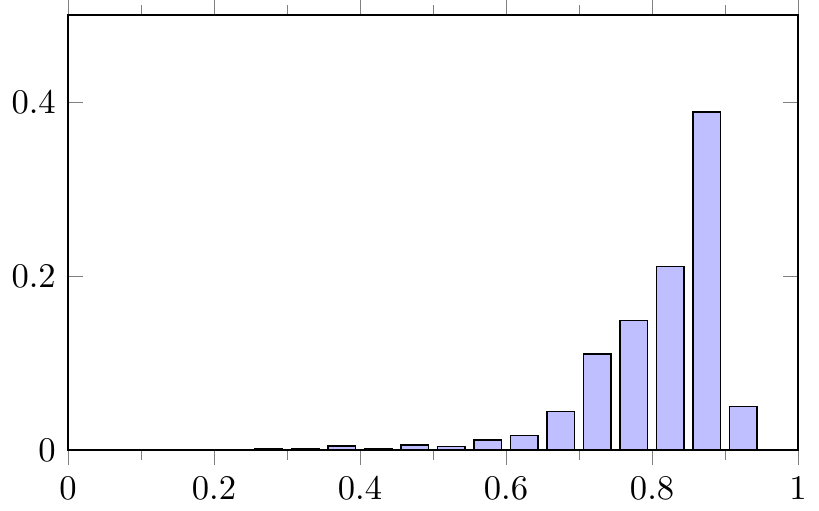}
        }
        \par\bigskip
        \subfigure{%
            \label{fig:gate-c}
            \includegraphics[width=0.3\textwidth]{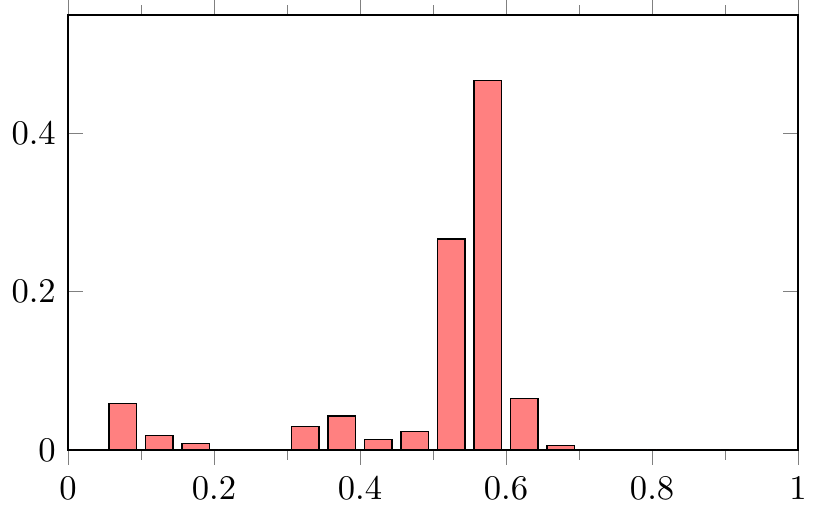}
        }
        \quad
        \subfigure{%
            \label{fig:gate-d}
            \includegraphics[width=0.3\textwidth]{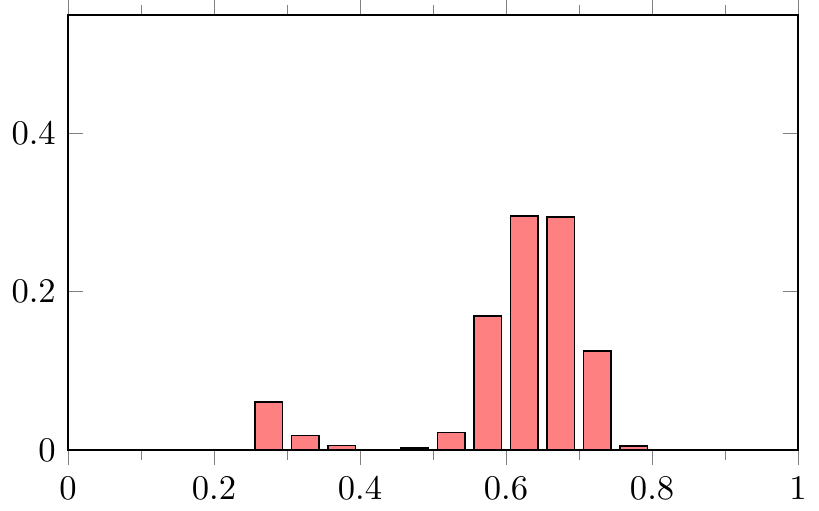}
        }
        \par\bigskip
        \subfigure{%
            \label{fig:gate-e}
            \includegraphics[width=0.3\textwidth]{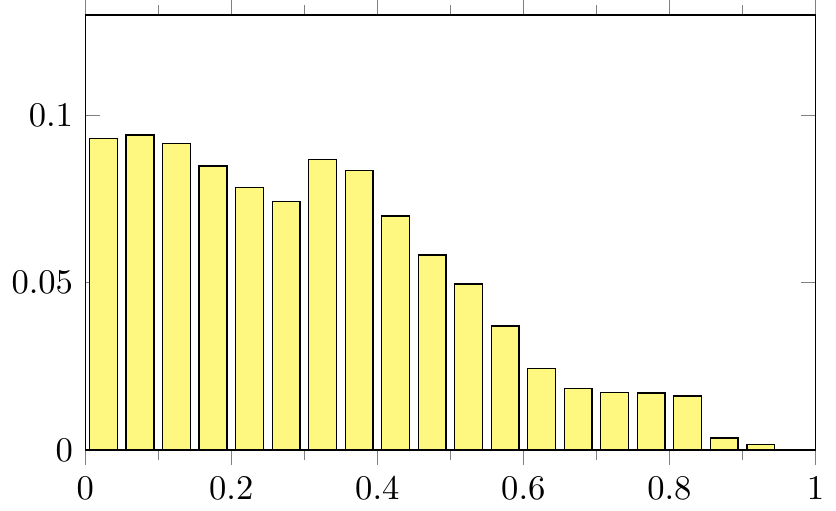}
        }
        \quad
        \subfigure{%
            \label{fig:gate-f}		
            \includegraphics[width=0.3\textwidth]{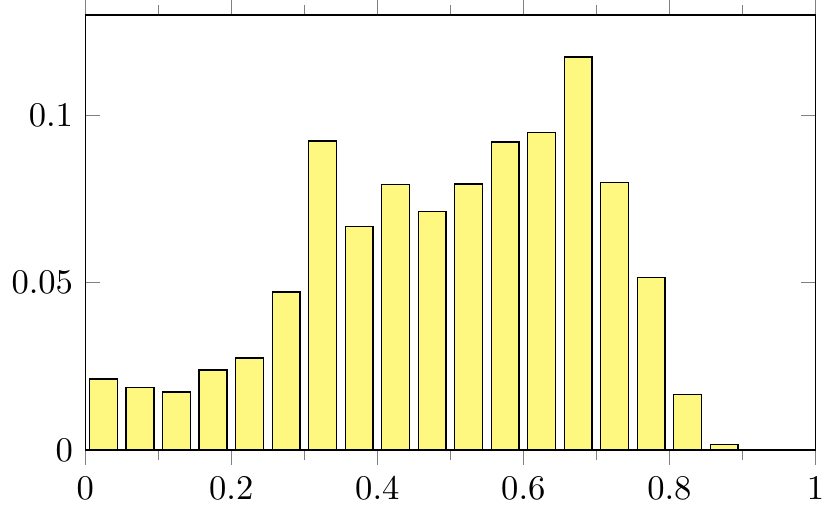}
        }
        \caption{
            (\textit{a}) $g^2_i$, (\textit{b}) $g^3_i$, (\textit{c}) $R(\mathbf{g}^2_\cdot)$,
            (\textit{d}) $R(\mathbf{g}^3_\cdot)$, (\textit{e}) $\vert g^2_i - o^1_i \vert$,
            (\textit{f}) $\vert g^3_i - o^2_i \vert$.
            (\textit{a}), (\textit{b}): Histograms of vertical forget gate values.
            (\textit{c}), (\textit{d}): Histograms of the ranges of vertical forget gate per time step.
            (\textit{e}), (\textit{f}): Histograms of the absolute difference between the previous output gate and the current vertical forget gate values.
        }
        \label{fig:gate-analysis}
    \end{figure}
    
    To inspect the effect of the additional forget gate, we investigate how the values of vertical forget gates are distributed.
    We sample 1,000 random sentences from the development set of the SNLI dataset, and use the 3-layer CAS-LSTM model trained on the SNLI dataset to compute gate values.
    
    If all values from a vertical forget gate $\mathbf{g}_t^l$ were to be 0, this would mean that the introduction of the additional forget gate is meaningless and the model would reduce to a plain stacked LSTM.
    On the contrary if all values were 1, meaning that the vertical forget gates were always \textit{open}, it would be impossible to say that the information is modulated effectively.
    
    Fig. \ref{fig:gate-a} and \ref{fig:gate-b} represent histograms of the vertical forget gate values from the second and the third layer.
    From the figures we can validate that the trained model does not fall into the degenerate case where vertical forget gates are ignored.
    Also the figures show that the values are right-skewed, which we conjecture to be a result of focusing more on a strong interaction between adjacent layers.
    
    To further verify that the gate values are diverse enough within each time step, we compute the distribution of the range of values per time step, $R(\mathbf{g}_t^l)=\max_i{g_{t,i}^l} - \min_i{{g}_{t,i}^l}$, where $\mathbf{g}_t^l=[g_{t,1}^l, \cdots, g_{t,d_l}^l]^\top$.
    We plot the histograms in Fig. \ref{fig:gate-c} and \ref{fig:gate-d}.
    From the figures we see that the vertical forget gate controls the amount of information flow effectively, making diverse decisions of retaining or discarding signals across dimensions.
    
    Finally, to investigate the argument presented in \S\ref{sec:caslstm} that the additional forget gate helps the previous output gate with reducing the burden of extracting all needed information, we inspect the distribution of the values from $\vert \mathbf{g}_t^l - \mathbf{o}_t^{l-1} \vert$.
    This distribution indicates how differently the vertical forget gate and the previous output gate select information from $\mathbf{c}_t^{l-1}$.
    From Fig. \ref{fig:gate-e} and \ref{fig:gate-f} we can see that the two gates make fairly different decisions, from which we demonstrate that the direct path between $\mathbf{c}_t^{l-1}$ and $\mathbf{c}_t^l$ enables a model to utilize signals overlooked by $\mathbf{o}_t^{l-1}$.
    
    \subsection{Model Variations}
    \label{exp:variations}
     
    In this subsection, we see the influence of each component of a model on performance by removing or replacing its components.
    the SNLI dataset is used for experiments, and the best performing configuration is used as a baseline for modifications.
    We consider the following variants: (\textit{i}) models with different $\bm\lambda$, (\textit{ii}) models without $\bm\lambda$, and (\textit{iii}) models that integrate lower contexts via peephole connections.
    
    Variant (\textit{iii}) calculates and applies the forget gate $\mathbf{g}_t^l$ which takes charge of integrating lower contexts via the equations below, following the work of \citet{zhang2016highway}:
    \begin{align}
    \begin{split}
    \mathbf{g}_t^l &= \sigma(\mathbf{W}_g^l \mathbf{h}_t^{l-1}
    + \mathbf{p}_{g_1}^l \odot \mathbf{c}_{t-1}^l + \mathbf{p}_{g_2}^l \odot \mathbf{c}_{t}^{l-1} + \mathbf{b}_g^l) \label{eq:peephole1}
    \end{split} \\
    \mathbf{c}_t^l &= \mathbf{i}_t^l \odot \tilde{\mathbf{c}}_t^l + \mathbf{f}_t^l \odot \mathbf{c}_{t-1}^l + \mathbf{g}_t^l \odot \mathbf{c}_t^{l-1}, \label{eq:peephole2}
    \end{align}
    where $\mathbf{p}_\cdot^l \in \mathbb{R}^{d_l}$ represent peephole weight vectors that take cell states into account.
    We can see that the computation formulae of $\mathbf{f}_t^l$ and $\mathbf{g}_t^l$ are not consistent, in that $\mathbf{h}_{t-1}^l$ does not participate in computing $\mathbf{g}_{t-1}^l$, and that the left and the below context are reflected in $\mathbf{g}_{t-1}^l$ only via element-wise multiplications which do not consider the interaction among dimensions.
    By contrast, ours uses the analogous formulae in calculating $\mathbf{f}_t^l$ and $\mathbf{g}_t^l$, considers $\mathbf{h}_{t-1}^{l}$ in calculating $\mathbf{g}_t^l$, and introduces the scaling factor $\bm{\lambda}$.
    
    Table \ref{table:variations} summarizes the results of model variants.
    From the results of \textit{baseline} and \textit{(i)}, we validate that the selection of $\bm\lambda$ does not significantly affect performance but introducing $\bm\lambda$ is beneficial (\textit{baseline vs. (ii)}) possibly due to its effect on normalizing information from multiple sources, as mentioned in \S\ref{sec:caslstm}.
    Also, from the comparison between  \textit{baseline} and \textit{(iii)}, we show that the proposed way of combining the left and the lower contexts leads to better modeling of sentence representations than that of \citet{zhang2016highway}.
    
    \begin{table}[t]
        \centering
        \begin{tabular}{l r r}
            \hline
            \bf{Model} & \bf{Acc. (\%)} & \bf{$\Delta$} \\
            \hline
            Bi-CAS-LSTM (\textit{baseline}) & 87.0 & \\
            \quad \textit{(i) Diverse $\bm\lambda$} & & \\
            \qquad \textit{(a) $\lambda_i=0.25$} & 86.8 & -0.2 \\
            \qquad \textit{(b) $\lambda_i=0.75$} &  86.8 & -0.2 \\
            \qquad \textit{(c) Trainable $\bm\lambda$} & 86.9 & -0.1 \\
            \quad \textit{(ii) No $\bm\lambda$} & 86.6 & -0.4 \\
            \quad \textit{(iii) Integration through peepholes} & 86.5 & -0.5 \\
            \hline
        \end{tabular}
        \caption{Results of model variants.}
        \label{table:variations}
    \end{table}
    
    \section{Conclusion}
    \label{sec:conclusion}
    In this paper, we proposed a method of stacking multiple LSTM layers for modeling sentences, dubbed CAS-LSTM.
    It uses not only hidden states but also cell states from the previous layer, for the purpose of controlling the vertical information flow in a more elaborate way.
    We evaluated the proposed method on various benchmark tasks: natural language inference, paraphrase identification, and sentiment classification.
    Our models outperformed plain LSTM-based models in all experiments and were competitive other state-of-the-art models.
    The proposed architecture can replace any stacked LSTM only under one weak restriction---the size of states should be identical across all layers.
    
    For future work we plan to apply the CAS-LSTM architecture beyond sentence modeling tasks.
    Various problems such as sequence labeling and language modeling might benefit from sophisticated modulation on context integration.
    Aggregating diverse contexts from sequential data, e.g. those from forward and backward reading of text, could also be an intriguing research direction.
    
    \section*{Acknowledgments}
    This work was supported by the National Research Foundation of Korea (NRF) grant funded by the Korea Government (MSIT) (NRF2016M3C4A7952587).
    
    \bibliography{acml19}
\end{document}